\documentclass[journal]{IEEEtran}

\usepackage[nocompress]{cite}
\usepackage[caption=false,font=footnotesize]{subfig}
\usepackage{listings}
\usepackage{xspace}
\usepackage{todonotes}
\usepackage{algorithmicx} % do not use the algorithm environment with
\usepackage{paralist}
\usepackage{units}
\usepackage{tikz}
\usetikzlibrary{calc,decorations,positioning,shapes.geometric}

\newcommand{\lang}    {Buzz\xspace}
\newcommand{\cmd}  [1]{\texttt{#1()}}
\newcommand{\fig}  [1]{\figurename~\ref{#1}}
\newcommand{\sect} [1]{Sec.~\ref{#1}}

\renewcommand{\paragraph}[1]{\noindent\textbf{#1}}

\lstdefinelanguage{buzz}{
  keywords={var, nil, if, else, function, return, for, while, and, or, not, swarm, stigmergy, neighbors, self, math},
  sensitive=true,
  comment=[l]\#,
  string=[b]",
  morestring=[b]',
  basicstyle=\scriptsize\ttfamily,
  keywordstyle=\color{black}\bfseries,
  identifierstyle=\color{black},
  commentstyle=\color{black!55}\itshape,
  showstringspaces=false
}
\begin{document}

% 
% Document data
% 
\title{\lang: An Extensible Programming Language for\\%
Self-Organizing Heterogeneous Robot Swarms}
\author{Carlo~Pinciroli, Adam~Lee-Brown, Giovanni~Beltrame%
\thanks{C.~Pinciroli and G.~Beltrame are with the Department of
  Computer and Software Engineering, \'Ecole Polytechnique de
  Montr\'eal, PO Box 6079, Succ. Centre-ville Montr\'eal, Qu\'ebec,
  Canada H3C 3A7. E-mail:
  \{carlo.pinciroli,giovanni.beltrame\}@polymtl.ca}%
\thanks{A.~Lee-Brown is with the School of Electrical and
  Electronic Engineering, Royal Melbourne Institute of Technology, 124
  La Trobe St, Melbourne VIC 3000, Australia. E-mail:
  s3434074@student.rmit.edu.au}%
}

\maketitle
% \IEEEpubid{0000--0000/00\$00.00~\copyright~2015 IEEE}

\begin{abstract}
  We present Buzz, a novel programming language for heterogeneous
  robot swarms. Buzz advocates a compositional approach, offering
  primitives to define swarm behaviors both from the perspective of
  the single robot and of the overall swarm.  Single-robot primitives
  include robot-specific instructions and manipulation of neighborhood
  data.  Swarm-based primitives allow for the dynamic management of
  robot teams, and for sharing information globally across the
  swarm. Self-organization stems from the completely decentralized
  mechanisms upon which the Buzz run-time platform is based. The
  language can be extended to add new primitives (thus supporting
  heterogeneous robot swarms), and its run-time platform is designed
  to be laid on top of other frameworks, such as Robot Operating
  System. We showcase the capabilities of Buzz by providing code
  examples, and analyze scalability and robustness of the run-time
  platform through realistic simulated experiments with representative
  swarm algorithms.
\end{abstract}


\section{Introduction}
\label{sec:introduction}

\IEEEPARstart{S}{warm} robotics systems~\cite{Beni2005} are envisioned
for large-scale application scenarios that require reliable, scalable,
and autonomous behaviors. Among the many hurdles towards real-world
deployment of swarm robotics systems, one of the most important is the
lack of dedicated tools, especially regarding
software~\cite{McLurkin2006}. In particular, one problem that has
received little attention in the literature is
\emph{programmability}. The current practice of swarm behavior
development typically focuses on individual behaviors and low-level
interactions~\cite{Brambilla2013}.  This approach forces developers to
constantly `reinvent the wheel' to fit well-known algorithms into new
applications, resulting in a slow and error-prone development process.

To promote code reuse, two general approaches are possible. The first
approach is the development of software libraries that encapsulate
certain algorithms. While this has the advantage of leveraging
pre-existing frameworks and tools, it falls short of screening the
user from unnecessary detail, such as non-trivial compilation
configuration or language-specific boilerplate.

The second approach is the creation of a domain-specific language
(DSL) exposing only the relevant aspects of the problem to solve. A
well-designed DSL shapes the way a system is conceived, by offering
powerful abstractions expressed through concise constructs.

In this paper, we argue that a DSL is an indispensable tool towards
the deployment of real-world robot swarms. The dynamics of robot
swarms are made complex by the presence of both spatial aspects (body
shape, sensing, actuation) and networks aspects (message loss,
volatile topology). Additionally, to render the production of
large-scale swarms affordable, the robots suffer from significant
limitations in terms of computational power, especially when compared
with robots designed to act alone.  Thus, a DSL for robot swarms has
the potential to act as a platform that
\begin{inparaenum}[\it (i)]
\item Filters the low-level details concerning space and networking,
  in an efficient, resource-aware fashion;
\item Offers a coherent abstraction of the system;
\item Acts as common platform for code reuse and benchmarking.
\end{inparaenum}
% Keep this here, it's to fix spacing at the bottom of the page
% \IEEEpubidadjcol

In this paper, we present Buzz, a novel DSL for robot swarms. To drive
the design of Buzz, we identified key requirements a successful
programming language for swarm robotics must meet.

First, as mentioned, the language must allow the programmer to work at
a suitable level of abstraction.  The complexity of concentrating on
individual robots and their interactions, i.e., a \emph{bottom-up}
approach, increases steeply with the size of the swarm. Conversely, a
purely \emph{top-down} approach, i.e., focused on the behaviour of the
swarm as a whole, might lack expressive power to fine-tune specific
robot behaviors. We believe that a language for robot swarms must
combine \emph{both} bottom-up and top-down primitives, allowing the
developer to pick the most comfortable level of abstraction to express
a swarm algorithm.

Second, the language must enable a compositional approach, by
providing predictable primitives that can be combined intuitively into
more complex algorithms and constructs.

Third, the language must prove generic enough to
\begin{inparaenum}[\it (i)]
\item express the most common algorithms for swarm coordination, such
  as flocking, task allocation, and collective decision making; and
\item support heterogeneous swarms.
\end{inparaenum}

Fourth, the run-time platform of the language must ensure acceptable
levels of scalability (for increasing swarm sizes) and robustness (in
case of temporary communication issues). Currently, the management of
the low-level aspects and corner cases concerning these issues
constitutes a sizable portion of the development process of swarm
behaviors. Alleviating this burden with a scalable and robust run-time
platform is crucial for real-world deployment of swarm behaviors.

The main contribution of this paper is the design and implementation
of Buzz, a programming language that meets the requirements discussed
above. Buzz is released as open-source software under the MIT
license. It can be downloaded at
\texttt{http://the.swarming.buzz/}.

The text is organized as follows. In \sect{sec:highlevel}, we present
the design principles we followed. In \sect{sec:primitives}, we
introduce the Buzz syntax. In \sect{sec:runtime} we illustrate the
run-time platform. In \sect{sec:examples} we report a number of
scripts meant to showcase the capabilities of Buzz, in particular its
conciseness and generality. In \sect{sec:experiments} we analyze the
scalability and robustness of the Buzz run-time. In \sect{sec:related}
we discuss related work, outlining similarities and differences of
Buzz with respect to existing languages and frameworks. Finally, in
\sect{sec:conclusions} we draw concluding remarks.

\section{Design Principles}
\label{sec:highlevel}

The design of Buzz was based on a set of high-level principles, that
are described in the following.

\paragraph{Discrete swarm, step-wise execution.}
A swarm is considered as a discrete collection of devices, each
running the Buzz virtual machine (BVM), and executing the same Buzz
script. Script execution proceeds independently on each robot in a
step-wise fashion. Each time-step is divided in five phases:
\begin{inparaenum}
\item Sensor readings are collected and stored in the BVM;
\item Incoming messages are collected and processed by the BVM;
\item A portion of the Buzz script is executed;
\item The messages in the BVM output queue are sent (as many as
  possible, according to the available payload size; see
  \sect{sec:runtime});
\item Actuator values are collected from the BVM state and applied.
\end{inparaenum}
The length of a time-step is constant over the life-time of the
script, although it does not need to be the same for every robot. If
the execution time of a step is shorter than the allotted time, the
robot sleeps for a period. If the execution is longer, a warning
message is printed on the screen and the execution proceeds.

\paragraph{Predictable and composable syntax.}
Buzz employs a simple, imperative syntax inspired by languages such
as Python, Lua, and JavaScript.  While this syntax ensures a short
learning curve, it also allows for predictable and composable
scripts. By \emph{predictable}, we mean that a programmer can easily
infer the results of the execution of a program by reading its
code. \emph{Composability} refers to the ability to structure a
complex program into simpler elements, such as functions, classes,
etc.\  that can later be used as modules.

\paragraph{Support for heterogeneous robots.}
Buzz is explicitly designed to support heterogeneous robot swarms. To
this aim, Buzz is conceived as an \emph{extension language}, that is,
a language that exports and enhances the capabilities of an already
existing system. This design choice allows one to stack the Buzz
run-time on top of well-known single-robot frameworks such as
ROS,\footnote{http://www.ros.org/}
OROCOS,\footnote{http://www.orocos.org/} or
YARP.\footnote{http://wiki.icub.org/yarp/} The Buzz syntax and
primitives are minimal, concise, and powerful; the programmer can add
new primitives and constructs that link to relevant features of the
underlying system.

\paragraph{Swarm-level abstraction.}
One of the novel aspects of Buzz is the ability to manage swarms of
robots. The concept of \emph{swarm} is a first-class language
object~\cite{Scott2006}. Swarms can be created and disbanded; new
swarms can be created as a result of an operation of intersection,
union, or difference between two pre-existing swarms. Each swarm can
be atomically assigned operations to perform. Swarms can be employed
to encode complex task coordination strategies.

\paragraph{Situated communication.}
Each robot is assumed to be equipped with a device capable of
\emph{situated communication}~\cite{Stoy2001}. Such device broadcasts
messages within a limited range and receives messages from neighboring
robots in direct, unobstructed line-of-sight. Upon receipt of a
message, a robot also detects the relative distance and angle of the
message sender. Situated communication has been used extensively in
swarm robotics to achieve global coordination in algorithms for, e.g.,
pattern formation~\cite{Spears2004}, flocking~\cite{Ferrante2014},
exploration~\cite{Pinciroli2015}, and task
allocation~\cite{Brutschy2014}.  This communication modality can be
achieved on-hardware with robots such as the
Kilobot~\cite{Rubenstein2012}, the e-puck~\cite{Mondada2006}, and the
marXbot~\cite{Bonani2010}, although with limited payload size, and a
dedicated module for low-cost 3D communication was proposed
in~\cite{DeSilva2014}. Alternatively, situated communication can be
satisfactorily realized through WiFi and GPS (for outdoor scenarios)
or tracking systems such as Vicon (for indoor scenarios).

\paragraph{Information sharing.}
Buzz provides two ways for robots to share information: virtual
stigmergy and neighbor queries. \emph{Virtual stigmergy} is a data
structure that allows the robots in a swarm to globally agree on the
values of a set of variables. From an abstract point of view, virtual
stigmergy is akin to a distributed key-value storage.  Among its many
potential applications, this structure may be employed to coordinate
task execution, e.g., mark the tasks that have already been completed
and/or share progress on the uncompleted ones.  \emph{Neighbor
  queries} allow a robot to inquire its direct neighbors on the value
of a certain variable. Upon receiving a request, the neighbors reply
with the current value of the variable; these replies are then
collected into a list by the robot who requested the data. Neighbor
queries are useful when robots must perform local data aggregation,
such as calculating averages~\cite{Jelasity2005} or constructing
spatial gradients.

\section{Language Definition}
\label{sec:primitives}

\subsection{Robot-wise operations and primitive types}
\label{sec:defprimitives}
The simplest operations available in Buzz are robot-wise
operations. These operations are executed by every robot
individually. Assignments, arithmetic operations, loops and
branches fall into this category. E.g.:
\begin{lstlisting}
# Assignment and arithmetic operation
a = 3 + 7
# Loop
i = 0; while(i < a) i = i + 1
# Branching
if(a == 10) i = 0
\end{lstlisting}

Buzz is a dynamically typed language that offers the following types:
\texttt{nil}, integer, floating-point, string, table, closure, swarm,
and virtual stigmergy. The \texttt{nil}, integer, floating-point, and
string types work analogously to other scripting languages such as
Python or Lua. The swarm and virtual stigmergy types are introduced in
\sect{sec:defswarms} and \ref{sec:defvstig}, respectively.

Tables are the only structured type available in Buzz, and they are
inspired by the analogous construct in
Lua~\cite{Ierusalimschy2005}. Tables can be used as arrays or
dictionaries.
\begin{lstlisting}
# Create an empty table
t = {}
# Index syntax, use as array
t[6] = 5
# Dot syntax, use as dictionary
t.b = 9
# Index syntax, use as dictionary
t["b"] = 10
\end{lstlisting}
Buzz also supports functions as first-class objects, implemented as
closures~\cite{Landin1964}. Two types of closures exist: native
closures and C closures. Native closures refer to functions defined
within a Buzz script, while C closures refer to C functions registered
into the BVM (see \sect{sec:implheterogeneous}).
\begin{lstlisting}
# Function definition
function f(a) { return a }
# Native closure assignment
n = f
# Function call using f. x is set to 9
x = f(9)
# Function call using n. x is set to 5
x = n(5)
# Native closures can be defined anonymously (a.k.a. lambdas)
l = function(a,b) { return a+b }
x = l(2,3) # x is set to 5
\end{lstlisting}
The combination of tables and closures allows for the creation of
class-like structures. Following the object-oriented programming
jargon, we refer to functions assigned to table elements as
\emph{methods}:
\begin{lstlisting}
# Create empty table
t = {}
# Add attribute to table
t.a = 4
# Add method to table
# 'self' refers to the table owning the method
t.m = function(p) { return self.a + p }
# Method call
x = t.m(6) # x is set to 10
\end{lstlisting}

\subsection{Swarm management}
\label{sec:defswarms}
Buzz lets the programmer subdivide the robots into multiple teams. In
Buzz parlance, each team is called a \emph{swarm}.  Buzz treats swarms
as first-class objects. To create a swarm, the programmer must provide
a unique identifier, which is known and shared by all the robots in
the created swarm. The returned value is a class-like structure of
type \texttt{swarm}.
\begin{lstlisting}
# Creation of a swarm with identifier 1
s = swarm.create(1)
\end{lstlisting}
Once a swarm is created, it is initially empty. To have robots join a
swarm, two methods are available: \cmd{select} and \cmd{join}. With
the former, the programmer can specify a condition, evaluated by each
robot individually, for joining a swarm; with the latter, a robot
joins a swarm unconditionally. To leave a swarm, Buzz offers the
analogous methods \cmd{unselect} and \cmd{leave}. A robot can check
whether it belongs to a swarm through the method \cmd{in}.
\begin{lstlisting}
# Join the swarm if the robot identifier (id) is even
# - 'id' is an internal symbol that refers to the
#   numeric id of the robot executing the script
# - % is the modulo operator
s.select(id % 2 == 0)
# Join the swarm unconditionally
s.join()
# Leave the swarm if the robot id is greater than 5
s.unselect(id > 5)
# Leave the swarm unconditionally
s.leave()
# Check whether a robot belongs to s
if(s.in()) { ... }
\end{lstlisting}

Once a swarm is created, it is possible to assign tasks to it through
the method \cmd{exec}. This method accepts a closure as
parameter.
\begin{lstlisting}
# Assigning a task to a swarm
s.exec(function() { ... })
\end{lstlisting}
Internally, the closure is executed as a \emph{swarm function
  call}. This call modality differs from normal closure calls in that
the current swarm id is pushed onto a dedicated \emph{swarm
  stack}. Upon return from a swarm function call, the swarm stack is
popped. When the swarm stack is non-empty, the \cmd{swarm.id} method
is defined. If called without arguments, it returns the swarm id at
the top of the swarm stack (i.e., the current swarm id); if passed an
integer argument \texttt{n} $>0$, it returns the \texttt{n}-th element
in the swarm stack. Besides making the \cmd{swarm.id} command
possible, the swarm stack is instrumental for neighbor operations (see
\sect{sec:implneighbors}).

The programmer can create new swarms that result from operations on
pre-existing swarms. Four such operations are available: intersection,
union, difference, and negation. The first three operations act on two
swarms:
\begin{lstlisting}
# a, b are swarms defined earlier in the script
# Create new swarm with robots belonging to both a and b
# The first argument is a unique swarm identifier
i = swarm.intersection(100, a, b)
# Create new swarm with robots belonging to a or b
u = swarm.union(101, a, b)
# Create new swarm with robots belonging to a and not to b
d = swarm.difference(102, a, b)
\end{lstlisting}
The fourth operation, negation, is encoded in the \cmd{others} method
and it creates a new swarm that contains all
the robots that do not belong to a given swarm:
\begin{lstlisting}
# Create a new swarm n as the negation of swarm s
n = s.others(103)
\end{lstlisting}

\subsection{Neighbor operations}
\label{sec:defneighbors}
Buzz offers a rich set of operations based on the neighborhood of a
robot. These operations include both spatial and communication
aspects. 

\paragraph{The \texttt{neighbors} structure.}
The entry point of all neighbor operations is the \texttt{neighbors}
structure. For each robot, this structure stores spatial information
on the neighbors within communication range. The structure is updated
at each time step. It is internally organized as a dictionary, in
which the index is the id of the neighbor and the data entry is a
tuple \texttt{(distance, azimuth, elevation)}.

\paragraph{Iteration, transformation, reduction.}
The \texttt{neighbors} structure admits three basic operations:
iteration, transformation, and reduction. Iteration, encoded in the
method \cmd{foreach}, allows the programmer to apply a function
without return value to each neighbor. For instance, to print the data
stored for each neighbor, one could write:
\begin{lstlisting}
# Iteration example (rid is the neighbor's id)
neighbors.foreach(
  function(rid, data) {
    print("robot ", rid, ": ",
          "distance  = ", data.distance, ", "
          "azimuth   = ", data.azimuth, ", "
          "elevation = ", data.elevation) })
\end{lstlisting}
Transformation, encoded in the method \cmd{map}, applies a function
with return value to each neighbor. The return value is the result of
an operation on the data associated to a neighbor. The end result of
mapping is a new \texttt{neighbors} structure, in which each neighbor
id is associated to the transformed data entries. For example, to
transform the neighbor data into cartesian coordinates, one could
proceed as follows:
\begin{lstlisting}
# Transformation example
cart = neighbors.map(
  function(rid, data) {
    var c = {}
    c.x = data.distance * math.cos(data.elevation) *
          math.cos(data.azimuth)
    c.y = data.distance * math.cos(data.elevation) *
          math.sin(data.azimuth)
    c.z = data.distance * math.sin(data.elevation)
    return c })
\end{lstlisting}
Reduction, encoded in the method \cmd{reduce}, applies a function to
each neighbor to produce a single result. For instance, this code sums
the cartesian vectors calculated in the previous example:
\begin{lstlisting}
# Reduction example. accum is a table
# with values x, y, and z, initialized to 0
result = cart.reduce(function(rid, data, accum) {
    accum.x = accum.x + data.x
    accum.y = accum.y + data.y
    accum.z = accum.z + data.z
    return accum
  }, {x=0, y=0, z=0})
\end{lstlisting}

\paragraph{Filtering.}
It is often useful to apply the presented operations to a subset of
neighbors. The \cmd{filter} method allows the programmer to apply a
predicate to each neighbor. 
% The predicate is a function that returns
% \texttt{true} if the neighbor must be kept, and \texttt{false}
% otherwise.
 The end result of the \cmd{filter} method is a new
\texttt{neighbors} structure storing the neighbors for which the
predicate (a function) was true.
% \footnote{In Buzz, \texttt{false} values correspond to
%   \texttt{nil} and integer 0. Anything else is considered
%   \texttt{true}.}
For instance, to filter the neighbors whose
distance is within \unit[1]{m} from a robot, one could write:
\begin{lstlisting}
# Filtering example
onemeter = neighbors.filter(function(rid, data) {
    # We assume the distance is expressed in centimeters
    return data.distance < 100 })
\end{lstlisting}
Another common necessity is filtering neighbors by their membership to
a swarm. The \cmd{kin} method returns a \texttt{neighbors} structure
that contains the robots that belong to the same top-of-the-stack
swarm as the current robot.  The \cmd{nonkin} method returns the
complementary structure. An example application for these methods is
offered in \sect{sec:examples}.

\paragraph{Communication.}
Another use for the \texttt{neighbors} structure is to exchange and
analyze local data. To make this possible, Buzz offers two methods:
\cmd{broadcast} and \cmd{listen}. The former allows the robot to
broadcast a \texttt{(key, value)} pair across its neighborhood. The
latter takes two inputs: the \texttt{key} to listen to, and a
\texttt{listener} function to execute upon receiving a value from a
neighbor. The BVM executes the \texttt{listener} whenever data is
received until the method \texttt{neighbors.ignore(key)} is called.
As an example, in \sect{sec:experiments} we report an experiment in
which a robot swarm forms a distance gradient from a robot acting as
source.

\subsection{Virtual stigmergy}
\label{sec:defvstig}
\emph{Virtual stigmergy} is a data structure that allows a swarm of
robots to share data globally. Essentially, virtual stigmergy works as
a distributed tuple space analogous to
Linda~\cite{Gelernter1985}. Three methods are available: \cmd{create},
\cmd{put}, and \cmd{get}. As the names suggest, \cmd{create} is a
method that creates a new virtual stigmergy structure, while \cmd{put}
and \cmd{get} access the structure, writing or reading \texttt{(key,
  value)} entries.
\begin{lstlisting}
# Create a new virtual stigmergy
# A unique id (1 here) must be passed
v = stigmergy.create(1)
# Write a (key,value) entry into the structure
v.put("a", 6)
# Read a value from the structure
x = v.get("a")
\end{lstlisting}
A virtual stigmergy structure can handle only a subset of the
primitives types: integer, floating-point, string, and table. These
types can be used either as keys or values.

The name \emph{virtual stigmergy} derives from the indirect,
environment-mediated communication of nest-building insects such as
ants and termites~\cite{Grasse1959}. The key idea in natural stigmergy
is that environmental modifications to organize the environment occur
in a step-wise fashion, whereby a modification performed by an
individual causes a behavioral response in another individual, without
the two individuals ever directly interacting. From the point of view
of the programmer, virtual stigmergy works as a virtual shared
environment: a robot modifies an entry, and this modification triggers
a reaction in another robot without the two having to interact
directly.

As shown in \sect{sec:examples}, virtual stigmergy is a powerful
concept that enables the implementation of a large class of swarm
behaviors.

\section{Run-Time Platform}
\label{sec:runtime}

\subsection{The Buzz virtual machine}
\label{sec:implbuzzvm}
\begin{figure}[t]
  \centering
  \begin{tikzpicture}[
    sensact/.style={inner sep=0,text width=3.75cm,minimum height=3ex,text centered,font=\footnotesize},
    module/.style={draw,inner sep=0,minimum height=1cm,anchor=south west,text centered,font=\footnotesize},
    arr/.style={very thick}
    ]
    % \draw[help lines] (0,0) grid (8cm,7cm);
    %
    % Sensor/actuator
    % 
    \node(sensor)  [sensact,anchor=south west] at (0,0) {Sensor data};
    \node(actuator)[sensact,anchor=south east] at (8,0) {Actuator data};
    \draw[very thick] (sensor.north west) -- (sensor.north east);
    \draw[very thick] (actuator.north west) -- (actuator.north east);
    % 
    % Nodes
    % 
    \node(inmsg) [text width=2   cm,module]at (0   cm,1   cm){Input Message FIFO};
    \node(heap)  [text width=2.5 cm,module]at (2.5 cm,1   cm){Heap};
    \node(outmsg)[text width=2   cm,module]at (6   cm,1   cm){Output Message FIFO};
    \node(vstig) [text width=2   cm,module]at (0.5 cm,2.5 cm){Virtual Stigmergy};
    \node(nbr)   [text width=2   cm,module]at (0.5 cm,4   cm){Neighbor Data};
    \node(swdata)[text width=2   cm,module]at (0.5 cm,5.5 cm){Swarm Data};
    \node(interp)[text width=3.75cm,module]at (3.75cm,3   cm){Interpreter};
    \node(astack)[text width=1.75cm,module]at (3.75cm,4.75cm){Activation Record Stack};
    \node(sstack)[text width=1.75cm,module]at (5.75cm,4.75cm){Swarm Stack};
    % 
    % Arrows
    %
    % sensor -> inmsg
    \draw[arr,->] (1cm,0.5cm) -- (1cm,1cm);
    % sensor -> heap
    \draw[arr,->] (3.125cm,0.5cm) -- (3.125cm,1cm);
    % outmsg -> actuator
    \draw[arr,->] (7cm,1cm) -- (7cm,0.5cm);
    % heap -> actuator
    \draw[arr,->] (4.625cm,1cm) -- (4.625cm,0.5cm);
    % inmsg -> swdata
    \draw[arr,->] (0.25cm,2cm) |- (0.5cm,6cm);
    % inmsg -> vstig
    \draw[arr,->] (0.25cm,3cm) -- (0.5cm,3cm);
    % inmsg -> nbr
    \draw[arr,->] (0.25cm,4.5cm) -- (0.5cm,4.5cm);
    % vstig <-> interp
    \draw[arr,<->] (2.5cm,3.25cm) -- (3.75cm,3.25cm);
    % vstig <-> heap
    \draw[arr,<->] (2.5cm,2.75cm) -|(2.75cm,2cm);
    % nbr <-> interp
    \draw[arr,<->] (2.5cm,4.66cm) -- (3.25cm,4.66cm) -- (3.25cm,3.5cm) -- (3.75cm,3.5cm);
    % nbr <-> heap
    \draw[arr,<-] (2.5cm,4.33cm) -| (3cm,3.35cm);
    \draw[arr,]   (3cm,3.15cm) -- (3cm,3.1cm);
    \draw[arr,->] (3cm,2.9cm) -- (3cm,2cm);
    % swdata <-> interp
    \draw[arr,<->] (2.5cm,5.83cm) -- (3.5cm,5.83cm) -- (3.5cm,3.75cm) -- (3.75cm,3.75cm);
    % astack <-> interp
    \draw[arr,<->] (4.625cm,4.75cm) -- (4.625cm,4cm);
    % sstack <-> interp
    \draw[arr,<->] (6.625cm,4.75cm) -- (6.625cm,4cm);
    % swdata -> outmsg
    \draw[arr,->] (2.5cm,6.16cm) -| (7.75cm,2cm);
    % interp -> actuators
    \draw[arr,->] (5.5cm,3cm) -- (5.5cm,0.5cm);
    % interp -> outmsg
    \draw[arr,->] (7cm,3cm) -- (7cm,2cm);
    % interp <-> heap
    \draw[arr,<->] (4.375cm,3cm) -- (4.375cm,2cm);
    % vstig -> outmsg
    \draw[arr] (2.5cm,3cm) -- (3.5cm,3cm) -- (3.5cm,2.5cm) -- (4.275cm,2.5cm);
    \draw[arr] (4.475cm,2.5cm) -- (5.4cm,2.5cm);
    \draw[arr,->] (5.6cm,2.5cm) -- (6.25cm,2.5cm) -- (6.25cm,2cm);
  \end{tikzpicture}
  \caption{The structure of the Buzz virtual machine.}
  \label{fig:buzzvm}
\end{figure}
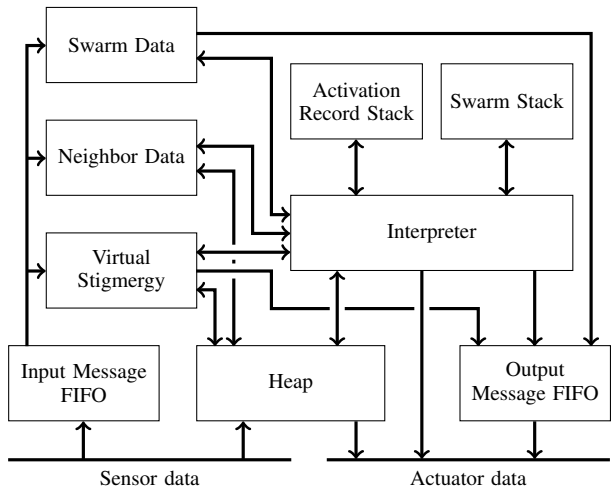

The run-time platform of Buzz is based on a custom, stack-based
virtual machine written entirely in C. The organization of the modules
composing the VM is depicted in \fig{fig:buzzvm}. The implementation
of a new VM is motivated by the design choices in Buzz. In particular,
the integration of swarm management and virtual stigmergy into a VM
for a dynamically-typed, extensible language forced us to find
dedicated solutions for data representation, stack/heap management,
and byte code encoding. The specifics on these aspects are purely
technical and go beyond the scope of this paper. A notable fact about
the BVM is its tiny size (\unit[12]{KB}) which fits most robots
currently in use for swarm robotics research.

At each time step, the BVM state is updated with the latest sensor
readings. Sensor readings are typically stored in the heap as data
structures (e.g., tables). Incoming messages are then inserted in the
BVM, which proceeds to unmarshal them and update the relevant modules.
Subsequently, the interpreter is called to execute a portion of the
script. At the end of this phase, the updated actuator data is read
from the BVM heap and part of the messages in the outbound queue are
sent by the underlying system.

\subsection{Code Compilation}
\label{sec:implcompilation}
The compilation of a Buzz script involves two tools: a compiler called
\texttt{buzzc} and an assembler/linker called \texttt{buzzasm}. The
former is a classical recursive descent parser that generates an
annotated object file in a single pass. The latter parses the object
file, performs the linking phase, and generates the byte code. The
compilation process is typically performed on the programmer's
machine, and the generated byte code is then uploaded on the robots.

Because Buzz is an extension language, it is likely for the compiler
to encounter unknown symbols. This typically occurs when robots with
different capabilities are employed. For instance, flying robots may
provide a \cmd{fly\_to} command for motion, while wheeled robots may
provide a \cmd{set\_wheels} command (see also
\sect{sec:implheterogeneous}).

By default, the compiler treats unknown symbols as global
symbols. Only at run-time, when a symbol is accessed, its actual value
is retrieved. In case the symbol is unknown at run-time, its value is
set to \texttt{nil}. This mechanism provides a flexible way to write
scripts that can work on robots with diverse capabilities. As a simple
example, to create a swarm that contains all the flying robots, it is
enough to write:
\begin{lstlisting}
aerial = swarm.create(1)
aerial.select(fly_to)
\end{lstlisting}

\subsection{Swarm management}
\label{sec:implswarm}
The management of swarm information is performed transparently by the
BVM.\@ Essentially, each robot maintains two data structures about
swarms: the first structure concerns the swarms of which the robot is
a member; the second stores data regarding neighboring robots. The
membership management mechanisms are loosely inspired by
CYCLON~\cite{Voulgaris2005}.

\paragraph{Membership management.}
Every time a \cmd{swarm.create} command is executed, the BVM stores
the identifier of the created swarm into a dedicated hash table, along
with a flag encoding whether the robot is a member of the swarm
(\texttt{1}) or not (\texttt{0}). Upon joining a swarm, the BVM sets
the flag corresponding to the swarm to \texttt{1} and queues a message
\texttt{<SWARM\_JOIN, robot\_id, swarm\_id>}. Analogously, when a
robot leaves a swarm, the BVM sets the corresponding flag to
\texttt{0} and queues a message \texttt{<SWARM\_LEAVE, robot\_id,
  swarm\_id>}. Because leaving and joining swarms is not a
particularly frequent operation, and motion constantly changes the
neighborhood of a robot, it is likely for a robot to encounter a
neighbor for which no information is available. To maintain
everybody's information up-to-date, the BVM periodically queues a
message \texttt{<SWARM\_LIST, robot\_id, swarm\_id\_list>} containing
the complete list of swarms to which the robot belongs. The frequency
of this message is chosen by the developer when configuring the BVM
installed on a robot.

\paragraph{Neighbor swarm data.}
The BVM stores the information in a hash map indexed by robot id. Each
element of the hash map is a \texttt{(swarm\_id\_list, age)} pair
where \texttt{swarm\_id\_list} corresponds the list of swarms of which
the robot is a member, and \texttt{age} is a counter of the time steps
since the last reception of a swarm update.  Upon receipt of a
swarm-related message (i.e., \texttt{SWARM\_JOIN},
\texttt{SWARM\_LEAVE}, \texttt{SWARM\_LIST}), the BVM updates the
information on a robot accordingly and zeroes the \texttt{age} of the
entry. This counter is employed to forget information on robots from
which no message has been received in a predefined period. When the
counter exceeds a threshold decided when configuring the BVM, the
information on the corresponding robot is removed from the
structure. This simple mechanism allows the robots to commit memory
storage on active, nearby robots while avoiding waste of resources on
unnecessary robots, such as out-of-range or damaged robots. In
addition, this mechanism prevents excessive memory usage when the
swarm size increases.

\paragraph{Message queue optimizations.}
To minimize the bandwidth required for swarm management, the BVM
performs a number of optimizations on the queue of swarm-related
outbound messages.
\begin{itemize}
\item If a \texttt{SWARM\_LIST} message is queued, the information
  contained in the message is more up-to-date than any other message
  in the queue. Thus, all the queued messages are removed and only the
  latest \texttt{SWARM\_LIST} message is kept.
\item If a \texttt{SWARM\_JOIN} message is queued, three situations
  can occur:\footnote{If an older \texttt{SWARM\_JOIN} message for the
    same swarm id is present, the new message is discarded.}
  \begin{inparaenum}[\it (i)]
  \item The queue does not contain any message regarding the same
    swarm id;
  \item The queue contains an older \texttt{SWARM\_LIST} message; or
  \item The queue contains an older \texttt{SWARM\_LEAVE} message for
    the same swarm id.
  \end{inparaenum}
  In the first case, the \texttt{SWARM\_JOIN} message is kept in the
  queue. In the second case, the \texttt{SWARM\_JOIN} message is
  dropped, the \texttt{SWARM\_LIST} message is kept in the queue, and
  the swarm id of the \texttt{SWARM\_JOIN} message is added to the
  list if not already present. In the third case, the
  \texttt{SWARM\_LEAVE} message is dropped and the
  \texttt{SWARM\_JOIN} message is kept in the queue.
\item If a \texttt{SWARM\_LEAVE} message is queued, three situations
  can occur:
  \begin{inparaenum}[\it (i)]
  \item The queue does not contain any message regarding the same
    swarm id;
  \item The queue contains an older \texttt{SWARM\_LIST} message; or
  \item The queue contains an older \texttt{SWARM\_JOIN} message for
    the same swarm id.
  \end{inparaenum}
  In the first case, the \texttt{SWARM\_LEAVE} message is kept in the
  queue. In the second case, the \texttt{SWARM\_LEAVE} message is
  dropped, the \texttt{SWARM\_LIST} message is kept in the queue, and
  the swarm id of the \texttt{SWARM\_LEAVE} message is removed from
  the list if present. In the third case, the \texttt{SWARM\_JOIN}
  message is dropped and the \texttt{SWARM\_LEAVE} message is kept in
  the queue.
\end{itemize}

\subsection{Neighbor operations}
\label{sec:implneighbors}
Neighbor operations involve the collection and manipulation of data
about nearby robots.

\paragraph{Neighbor data handling.}
Neighbor information is stored in a sparse array indexed by the robot
id of a neighbor. Each entry is a tuple containing the id of the
neighbor, its distance, azimuth, and elevation angles expressed
with respect to the robot's frame of reference. 

\paragraph{Neighbor data collection.}
Data collection involves the use of situated communication devices, as
discussed in \sect{sec:highlevel}. At each time step, a robot
broadcasts a message \texttt{<robot\_id>}. Upon receiving this
message, neighboring robots use their situated communication devices to
detect the distance, azimuth and elevation of the sending robot. The
\texttt{neighbors} structure is cleared and reconstructed at each time
step, to ensure that the data is constantly up-to-date. This operation
is not expensive, because typically the number of neighbors of a robot
is lower than 10.

\paragraph{Neighbor queries.}
Neighbor queries allow robots to inquire and share information on the
value of a specific symbol. When a robot executes the command
\texttt{neighbors.listen(key, listener)}, the BVM stores the passed
\texttt{listener} in a map indexed by \texttt{key}. Whenever the robot
processes a message identified by \texttt{key}, the BVM executes the
corresponding \texttt{listener}. The command
\texttt{neighbors.ignore(key)} removes \texttt{listener} from the
map. The method \texttt{neighbors.broadcast(key, value)} queues a
message \texttt{<key, value>}. If a message with the same \texttt{key}
is already present in the output queue, the BVM keeps the most recent
one.

\subsection{Virtual stigmergy}
\label{sec:implvstig}
Virtual stigmergy structures are stored by the BVM in a sparse array
indexed by the id provided in the Buzz script. Each entry of the
sparse array is a separate virtual stigmergy structure. Internally, a
virtual stigmergy structure is a hash map that stores tuples
\texttt{(key, value, timestamp, robot\_id)} where \texttt{key} is a
Buzz primitive type that identifies the entry, \texttt{value} is a
primitive type that contains the value of the entry,
\texttt{timestamp} is a Lamport clock~\cite{Lamport1978} used to
impose a temporal ordering on the entry updates, and
\texttt{robot\_id} is the id of the robot that last changed the entry.

\paragraph{Writing into a virtual stigmergy structure.}
When a robot writes a new value into a virtual stigmergy structure,
the BVM first creates or updates the local entry in the hash
map. Subsequently, the BVM queues a message \texttt{<VSTIG\_PUT,
  vstig\_id, key, value, timestamp, robot\_id>}. Nearby robots, upon
receipt of the message, check whether its \texttt{timestamp} is higher
than the locally known one. If this is the case, the robots propagate
the message. Otherwise, they ignore it. It might happen that two
robots advertise an update on the same \texttt{key} with the same
\texttt{timestamp}. When this happens, a user-defined conflict-solver
function is called. This function is given the conflicting entries as
parameters, and must return an entry as result. The resulting entry
can be either picked as-is from the input ones, or be a newly created
one. To help resolve conflicts, each input entry also contains a
\texttt{robot\_id} field storing the robot id that generated the
entry. If no conflict solver is specified by the user, the entry with
the highest robot id is propagated. A further user-defined function is
executed by the robot that lost the conflict. By default, this
function does nothing; however, in some applications a robot might
need to react, e.g.\ retry sending its update. An example usage of
these functions is reported in \sect{sec:example_decision}.

\paragraph{Reading from a virtual stigmergy structure.}
When a robot \emph{R1} reads a value from a virtual stigmergy
structure, the locally known value is returned by the
BVM.\@ Subsequently, the BVM queues a message \texttt{<VSTIG\_GET,
  vstig\_id, key, value1, timestamp1, robot\_id1>} to inquire nearby
robots on whether the local entry is up-to-date or not. Upon receiving
this message, a robot \emph{R2} checks its internal data structure. If
\emph{R2} knows more up-to-date information, it replies with a message
\texttt{<VSTIG\_PUT, vstig\_id, key, value2, timestamp2, robot\_id2>}
containing its local version of the entry. If, instead, \emph{R2}
possesses older information, its BVM updates the entry and then
broadcasts a message \texttt{<VSTIG\_PUT, vstig\_id, key, value1,
  timestamp1, robot\_id1>}. This mechanism allows robots to
automatically update information after temporary disconnections or
when random message loss occurs. 

\paragraph{Message queue optimizations.}
The fact that messages are sent only when an entry is written or read
ensures that maximum resources are concentrated on `hot' data. This
allows the robots to avoid expensive updates of the entire tuple
space. In this way, a virtual stigmergy structure can grow in size if
necessary, knowing that minimal overhead is necessary to keep `hot'
data up-to-date. However, the fact that each access to a virtual
stigmergy structure entails the production of a message can quickly
fill the message queue. To avoid this problem, when a message is
queued, the BVM checks for existing messages in the outbound queue
that refer to the same entry in the same virtual stigmergy structure,
and only keeps the most up-to-date message.

\subsection{Integrating and Extending Buzz}
\label{sec:implheterogeneous}
As an extension language, Buzz provides the necessary mechanisms to
add new commands and add data structures that export parts of an
underlying system (e.g. ROS~\cite{Quigley2009}).

\paragraph{Adding new data structures.}
Sensor readings are a common aspect that must be integrated with
Buzz. Typically, this is realized by adding dedicated data structures
(e.g., tables) to the BVM.\@ For instance, the e-puck is equipped with
8 proximity sensors placed in a ring around the robot body. The C code
to create a Buzz table that contains these values is as follows:
\begin{lstlisting}[language=c]
/* The Buzz virtual machine, assumed already initialized */
buzzvm_t vm;
/* The proximity readings, assumed already filled */
int prox[8];
/* Create a new Buzz table for the proximity readings */
buzzvm_pusht(vm);
buzzobj_t pt = buzzvm_stack_at(vm, 1);
buzzvm_pop(vm);
/* Store the proximity readings in the table */
for(int i = 0; i < 8; ++i) {
  buzzvm_push(pt);           /* push table */
  buzzvm_pushi(vm, i);       /* push table index */
  buzzvm_pushi(vm, prox[i]); /* push prox reading */
  buzzvm_tput(vm);           /* store in table */
}
/* Store the table as the global symbol "prox" */
buzzvm_pushs(vm, buzzvm_string_register(vm, "prox"));
buzzvm_push(pt);
buzzvm_gstore(pt);
\end{lstlisting}
In Buzz, one could then write the following:
\begin{lstlisting}
if(prox[0] > 100 or prox[7] > 100) {
  # obstacle in front, turn around
}
\end{lstlisting}

\paragraph{Adding new commands.}
Adding new commands allows a programmer to extend Buzz with new
capabilities. External C functions are integrated with Buzz as C
closures. 
% This is the signature of a C function for Buzz:
% \begin{lstlisting}[language=c]
% int f(buzzvm_t vm) {
%   /* Do something */
%   /* The function returns nothing to Buzz */
%   /* return buzzvm_ret0(vm); */
%   /* The function returns the stack-top value to Buzz */
%   /* return buzzvm_ret1(vm); */
% }
% \end{lstlisting}
For instance, the code to integrate a function that sets the e-puck
wheel speeds is as follows (error checking is omitted for brevity):
\begin{lstlisting}[language=c]
int set_wheels(buzzvm_t vm) {
  /* Read arguments */
  buzzvm_lload(vm, 1);
  buzzvm_lload(vm, 2);
  int leftwheel = buzzvm_stack_at(vm, 2);
  int rightwheel = buzzvm_stack_at(vm, 1);
  /* Set wheel speed using low-level e-puck library */
  ...
  /* Return no value to Buzz */
  return buzzvm_ret0(vm);
}
/* Register function as global symbol */
buzzvm_pushs(vm, buzzvm_string_register(vm, "set_wheels"));
buzzvm_pushcc(vm, buzzvm_function_register(vm, set_wheels));
buzzvm_gstore(vm);
\end{lstlisting}
As a result, in Buzz one could write:
\begin{lstlisting}
# Set e-puck wheel speeds
set_wheels(10.0, 5.0)
\end{lstlisting}

% \paragraph{User data.}
% Through the presented commands, any type of C function or data
% structure can be integrated with Buzz. However, C++ frameworks such as
% ROS require to bind Buzz functions to class methods. For this, an
% additional Buzz primitive type is necessary: the \emph{user
%   data}. This type stores a \texttt{void*} pointer to a user-defined
% C/C++ entity. For instance, to map a C++ method call to a Buzz
% function one could proceed as shown below:
% \begin{lstlisting}[language=c++]
% // Class and object definition
% class C {
%   public:
%     void method();
% } c;

% // Register pointer to 'c' as global symbol
% buzzvm_pushs(vm, buzzvm_register_string(vm, "point_to_c"));
% buzzvm_pushuserdata(vm, &c);
% buzzvm_gstore(vm);

% // Registered C function (glue code)
% int call_method(buzzvm_t vm) {
%   // Get global symbol "point_to_c"
%   buzzvm_pushs(vm, buzzvm_register_string(vm, "point_to_c"));
%   buzzobj_t p = buzzvm_stack_at(vm, 1);
%   // Call class method
%   reinterpret_cast<C*>(p->u.value)->method();
%   // Return no value to Buzz
%   return buzzvm_ret0(vm);
% }

% // Register C function into Buzz
% buzzvm_pushs(vm, buzzvm_register_string(vm, "method"));
% buzzvm_pushcc(vm, buzzvm_register_function(vm, call_method));
% buzzvm_gstore(vm);
% \end{lstlisting}
% In Buzz, one can simply call the method as \texttt{method();}.
% % \begin{lstlisting}
% % method();
% % \end{lstlisting}

\paragraph{Calling Buzz functions from C.}
It is sometimes necessary to call a (native or C) closure from C
code. This happens, for instance, when the execution loop is managed
outside Buzz. In this case, a Buzz script is typically organized in
functions such as \cmd{init}, \cmd{step}, and \cmd{destroy}. These
functions are called by the underlying system when necessary. As a
simple example, let us take the increment function:
\begin{lstlisting}
# Buzz function that accepts a single argument
function inc(x) { return x + 1 }
\end{lstlisting}
To call it from C code, one needs to push its argument on the stack
(an integer, in this example) and then call the function
\cmd{buzzvm\_function\_call}:
\begin{lstlisting}[language=c]
/* Push the function argument on the stack */
buzzvm_pushi(vm, 5);
/* Call the function with:
 * 1. The VM data
 * 2. The function name
 * 3. The number of parameters passed
 */
buzzvm_function_call(vm, "inc", 1);
\end{lstlisting}

\section{Examples}
\label{sec:examples}
The objective of this section is to showcase Buzz, by providing
examples of common swarm algorithms for motion and task
coordination. These examples show how to build complex swarm behaviors
starting from simpler primitives, and demonstrate the generality and
composability of Buzz. The examples are also designed to suggest how a
library of reusable swarm behaviors could be constructed using Buzz.
The scripts were tested with ARGoS~\cite{Pinciroli2012}, an accurate,
physics-based simulator that includes models for
Spiri\footnote{http://www.pleiades.ca}, a commercial quad-rotor robot.

\subsection{Motion and Spatial Coordination}
\label{sec:example_motion}
% Motion is a fundamental capability in robot swarms. 
In heterogeneous swarms, the presence of robots with diverse motion
means can enhance the capabilities of the swarm~\cite{Dorigo2013}.
Buzz does not offer native motion primitives, leaving the designer
with the freedom to set ones that are most suitable for each robot. In
this section, we assume that every robot is endowed with a primitive
\cmd{goto}, which takes a 2D direction as input in the form of a table
\texttt{\{x,y\}}. The direction is then transformed into low-level
actuation accordingly to the specific motion means of a robot (e.g.,
wheels, propellers).
\begin{figure*}[!t]
  \centering
  \subfloat[Hexagonal pattern formation.]{
    \includegraphics[width=.8\columnwidth]{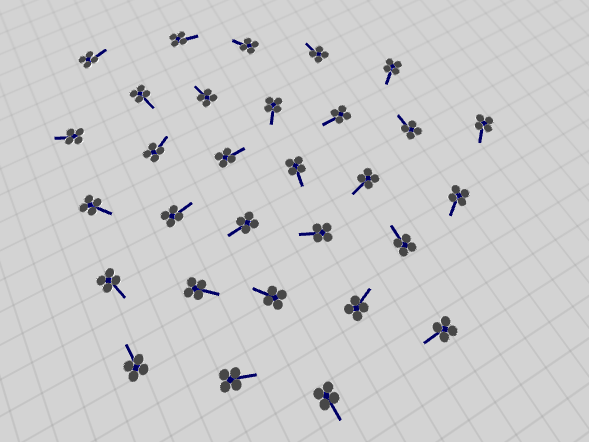}\label{fig:hexagon}
  }
  \hfill
  \subfloat[Segregation.]{
    \includegraphics[width=.8\columnwidth]{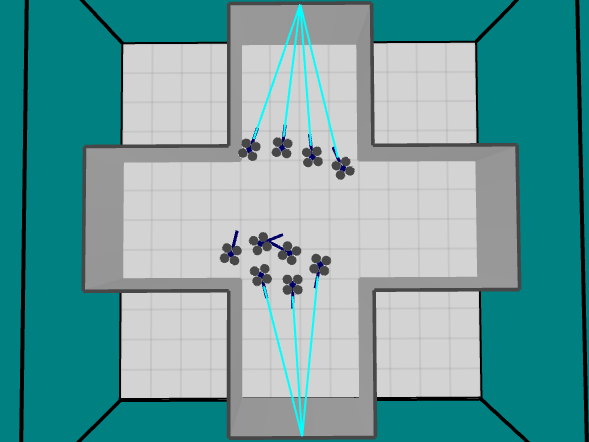}\label{fig:segregation}
  }
  \caption{ARGoS screenshots of the swarm behaviors presented in
    \sect{sec:example_motion}.}\label{fig:motion}
\end{figure*}

Pattern formation~\cite{Spears2004} is a basic swarm behavior that has
been employed in several applications including
flocking~\cite{Ferrante2014} and area
coverage~\cite{Pinciroli2015}. One of the most common approaches, and
the one we show here, is to implement pattern formation through
virtual physics~\cite{Spears2004}. Essentially, every robot is
considered as a charged particle immersed in a potential field, which
is `virtual' because it is calculated from neighbor positioning
data. The direction vector of each robot is calculated as the weighted
sum of the virtual forces that derive from the interaction of a robot
with its neighbors. We employ the following formula~\cite{Spears2004}
to calculate the magnitude of the virtual force $\mathbf{f}_n$ due to
a neighbor $n$:
$$
\left|\mathbf{f}_n\right| = -\frac{\epsilon}{d_n}
\left(\left(\frac{\delta}{d_n}\right)^4 -
  \left(\frac{\delta}{d_n}\right)^2\right),
$$
where $d_n$ is the current distance between the robot and its neighbor
$n$, $\delta$ is the target distance, and $\epsilon$ is a gain. The
direction vector a robot must follow is calculated as the average of
the vectors of the virtual forces $\mathbf{f}_n$:
$$
\mathbf{f} = \frac{1}{N} \sum_{n=1}^N \mathbf{f}_n.
$$

The \texttt{neighbors} structure provides a natural way to implement this
behavior. First, one must implement the function that calculates
$|\mathbf{f}_n|$:
\begin{lstlisting}
# Virtual force parameters (manually fitted)
DELTA   = 50.
EPSILON = 2700.
# Virtual force magnitude
function force_mag(dist, delta, epsilon) {
  return -(epsilon / dist) *
         ((delta / dist)^4 - (delta / dist)^2)
}
\end{lstlisting}
Next, we need a function to calculate $\mathbf{f}_n$ from the
positional information of a neighbor. This function is then used in
\cmd{neighbor.reduce} to calculate $\mathbf{f}$:
\begin{lstlisting}
# Virtual force accumulator
function force_sum(rid, data, accum) {
  var fm = force_mag(data.distance, DELTA, EPSILON)
  accum.x = accum.x + fm * math.cos(data.azimuth)
  accum.y = accum.y + fm * math.sin(data.azimuth)
  return accum
}
# Calculates the direction vector
function direction() {
  var dir = neighbors.reduce(force_sum, {x=0,y=0})
  dir.x = dir.x / neighbors.count()
  dir.y = dir.y / neighbors.count()
  return dir
}
# Executed at each step
# The loop is assumed managed outside Buzz
function step() {
  goto(direction())
}
\end{lstlisting}
\fig{fig:hexagon} reports an ARGoS screenshot depicting the structure
achieved by the robots after 30 seconds of simulated time.

\subsection{Collective Decision-Making}
\label{sec:example_decision}
Collective decision-making is an important aspect in multi-robot
coordination. In this section, we concentrate on one of the simplest
and widespread behaviors in this class---the barrier. A \emph{barrier}
is a synchronization mechanism that allows a swarm to wait until a
sufficient number of robots are ready to proceed.  Quorum sensing has
been proposed as a way to achieve this
behavior~\cite{Nagpal2004}. Quorum sensing refers to the progressive
build-up of a chemical in cell colonies. When the cells detect that
the concentration of such chemical exceeds a threshold, a new behavior
is triggered.

Virtual stigmergy offers a mean to implement quorum sensing. Every
time a robot wants to mark itself as ready, it puts a pair
\texttt{(id, 1)} in the virtual stigmergy structure, where \texttt{id}
is a robot's numeric id. The \cmd{size} method returns the number of
tuples in the virtual stigmergy, which here corresponds to the count
of ready robots.  When this count reaches the swarm size (or another
application-dependent threshold value), the robots collectively
trigger the next phase.
\begin{lstlisting}
# A numeric id for the barrier virtual stigmergy
BARRIER_VSTIG = 1
# Function to mark a robot ready
function barrier_set() {
  # Create a vstig
  barrier = stigmergy.create(BARRIER_VSTIG)
}
# Function to mark a robot ready
function barrier_ready() {
  barrier.put(id, 1)
}
# Function to wait for everybody to be ready
function barrier_wait(threshold) {
  while(barrier.size() < threshold) barrier.get(id);
}
\end{lstlisting}
An example usage of the barrier is presented in
\sect{sec:example_separation}.

\subsection{Separation into Multiple Swarms}
\label{sec:example_separation}
Segregation is a basic motion behavior whereby a group of robots
divides into two or more subgroups. A possible way to implement this
behavior in Buzz consists of defining two swarms, and use pattern
formation to impose a short distance between kin robots, and a long
one between non-kin ones. The following modifications to the pattern
formation algorithm in \sect{sec:example_motion} capture the
interaction among the different swarms:
\begin{lstlisting}
# Virtual force parameters (manually fitted)
DELTA_KIN      = 50.
EPSILON_KIN    = 2700.
DELTA_NONKIN   = 150.
EPSILON_NONKIN = 8000.
# Virtual force accumulator for kin robots
# Same as force_sum, with _KIN constants
function force_sum_kin(rid, data, accum) { ... }
# Virtual force accumulator for non-kin robots
# Same as force_sum, with _NONKIN constants
function force_sum_nonkin(rid, data, accum) { ... }
# Calculates the direction vector
function direction() {
  var dir
  dir = neighbors.kin().reduce(force_sum_kin, {x=0,y=0})
  dir = neighbors.nonkin().reduce(force_sum_nonkin, dir)
  dir.x = dir.x / neighbors.count()
  dir.y = dir.y / neighbors.count()
  return dir
}
\end{lstlisting}
The decision on which group a robot belongs to depends on the
application. In the following, we concentrate on the case in which two
different targets are present in the environment. Each target is
marked by a colored light---one red, one blue. The Spiri is equipped
with a frontal camera that can detect a target and its
color. We assume that not all of the robots are capable of detecting
the target, due to obstructions or sensor range limitations. The
uninformed robots must rely on the information shared by the informed
robots to pick the closest target. A possible solution to achieve this
is employing a virtual stigmergy structure, in which each robot
advertises its distance to the closest target and its color:
\begin{lstlisting}
NUM_ROBOTS   = 10
MAX_DISTANCE = 10000 # 100 meters
TARGET_VSTIG = 2
# Create virtual stigmergy
targetvstig = stigmergy.create(TARGET_VSTIG)
# Get target data
var mytargetdata = {}
if(camera.targetdata)
  # Can see the target directly
  mytargetdata = camera.targetdata
  targetvstig.put(id, mytargetdata)
  targetfound = 1
else {
  # Can't see the target directly
  mytargetdata.dist    = MAX_DISTANCE
  mytargetdata.color   = nil
  mytargetdata.closest = nil
  targetfound = nil
}
# Keep monitoring neighbors until everybody
# advertises a distance
while(not targetfound) {
  targetfound = 1
  mytargetdata = neighbors.reduce(
    function(rid, rdata, accum) {
      var d = targetvstig.get(rid)
      if(d != nil) {
        if(d.dist < DISTANCE_MAX) {
          if(accum.dist > rdata.distance + d.dist) {
            accum.dist    = rdata.distance + d.dist
            accum.closest = rid
            accum.color   = d.color
          }
        }
        else targetfound = nil
      }
      else targetfound = nil
      return accum
    }, mytargetdata)
  # Advertise choice
  targetvstig.put(id, mytargetdata)
  # Neighbors done?
  if(targetfound) barrier_ready()
}
# Wait for others to finish
barrier_wait(NUM_ROBOTS);
# When we get here, everybody has picked a target
\end{lstlisting}

Once the choice is done and the barrier is overcome, the robots form
two swarms and move accordingly:
\begin{lstlisting}
sred = swarm.create(COLOR_RED)
sred.select(mytargetdata.color == COLOR_RED)
sblue = swarm.create(COLOR_BLUE)
sblue.select(mytargetdata.color == COLOR_BLUE)
while(1) {
  sred.exec(function() { goto(direction()) })
  sblue.exec(function() { goto(direction()) })
}
\end{lstlisting}

\fig{fig:segregation} reports an ARGoS screenshot depicting the
structure achieved by the robots after 2 minutes of simulated time.

\section{Experimental Evaluation}
\label{sec:experiments}
In this section, we analyze the scalability and robustness of the Buzz
run-time. We focus our analysis on two subsystems, virtual
stigmergy and neighbor queries, because of their key role in swarm
coordination. The evaluation is conducted through simulations in ARGoS
with the ground-based marXbot robot~\cite{Bonani2010}.

\paragraph{Experimental setup.}
Our experimental setup consists of a square arena of side $L$ in which
$N$ robots are scattered. The coordinates $(x,y)$ of each robot are
chosen uniformly from $\mathcal{U}(-L/2,L/2)$.\footnote{When a
  coordinate choice causes physical overlap with already placed
  robots, a new coordinate is picked until no overlap occurs.} We
define the robot density $D$ as the ratio between the area occupied by
all the robots and the total area of the arena. To ensure comparable
conditions across different choices of $N$, we keep the density
constant ($D = 0.1$) and calculate $L$ with:
$$
D = \frac{N \pi R^2}{L^2} \Rightarrow L = \sqrt{\frac{N \pi R^2}{D}},
$$
where $R = \unit[8.5]{cm}$ is the radius of a marXbot. We focus our
analysis on two parameters that directly affect the properties that we
intend to analyze:
\begin{inparaenum}[\it (i)]
\item The number of robots $N$, which impacts scalability; and
\item The message dropping probability $P$, which affects robustness
  and accounts for an important, unavoidable phenomenon that
  influences the efficiency of current devices for situated
  communication.
\end{inparaenum}
For $N$, we chose $\{10, 100, 1000\}$; for $P$, we chose
$\{0, 0.25, 0.5, 0.75, 0.95\}$.  Each experimental configuration
$\langle N,P \rangle$ was tested 100 times.

\begin{figure}[t]
  \includegraphics[width=\columnwidth]{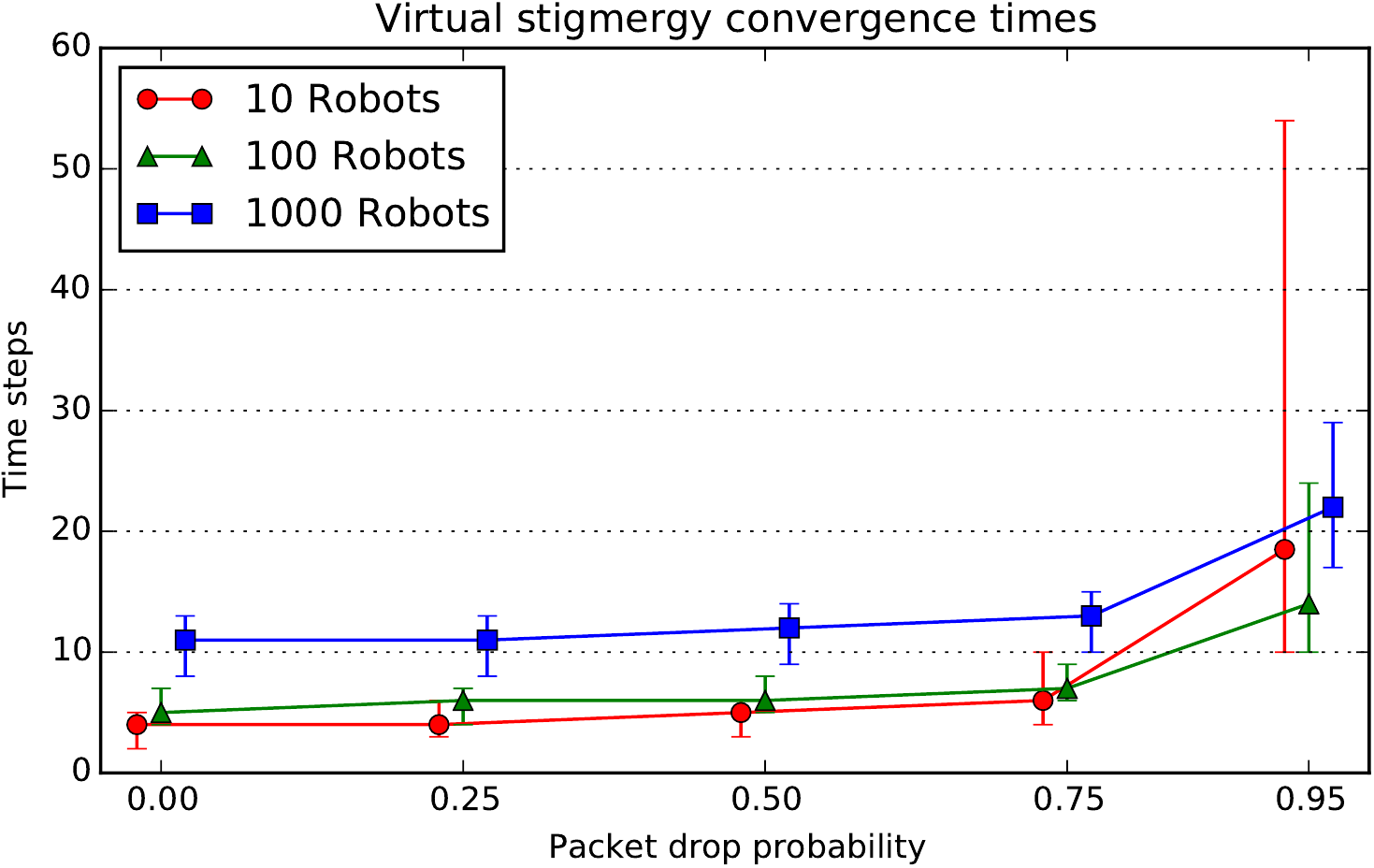}
  \begin{lstlisting}
# Executed at init time
function init() {
  # Create a vstig
  VSKEY = 1
  vs = stigmergy.create(1)
  # Set onconflict manager
  vs.onconflict(function(k,l,r) {
    # Return local value if
    # - Remote value is smaller than local, OR
    # - Values are equal, robot of remote record is
    #   smaller than local one
    if(r.data < l.data or
      (r.data == l.data and
       r.robot < l.robot)) {
      return l
    }
    # Otherwise return remote value
    else return r
  })
  # Initialize vstig
  vs_value = id
  vs.put(VSKEY, vs_value)
}

# Executed at each time step
function step() {
  # Get current value
  vs_value = vs.get(VSKEY)
}
  \end{lstlisting}
  \caption{Virtual stigmergy performance assessment. The plot reports
    the median, max, and min values of the distributions obtained for
    each experimental configuration $\langle N, P\rangle$. The markers
    are slightly offset to make them visible.}
  \label{fig:vs_exp}
\end{figure}

\paragraph{Virtual stigmergy.}
To analyze the efficiency of virtual stigmergy, we devised an
experiment in which the robots must agree on the highest robot id
across the swarm. This experiment is representative of a wide class of
situations in which a robot swarm must agree on the maximum or minimum
value of a quantity (e.g., sensor reading). As performance measure, we
employed the number of time steps necessary to reach global
consensus. We report in \fig{fig:vs_exp} the script we executed and
the data distribution we obtained. Our results indicate that, up to
$P = 0.75$, the number of time steps necessary to reach consensus is
affected weakly by $N$, and practically unaffected by
$P$. Interestingly, for $\langle N=1000,P=0.75 \rangle$, consensus is
reached in at most 15 time steps (a time step corresponds to
$\unit[0.1]{s}$ in our simulations). This positive result can be
explained by noting that, with $P=0.75$, 3 messages out of 4 are lost;
however, the uniform distribution of the robots ensures that, on
average, each robot has more than 3 neighbors. Thus, messages can
still flow throughout the network. The effect of packet dropping
is apparent for very pessimistic values ($P=0.95$).

\begin{figure}[t]
  \includegraphics[width=\columnwidth]{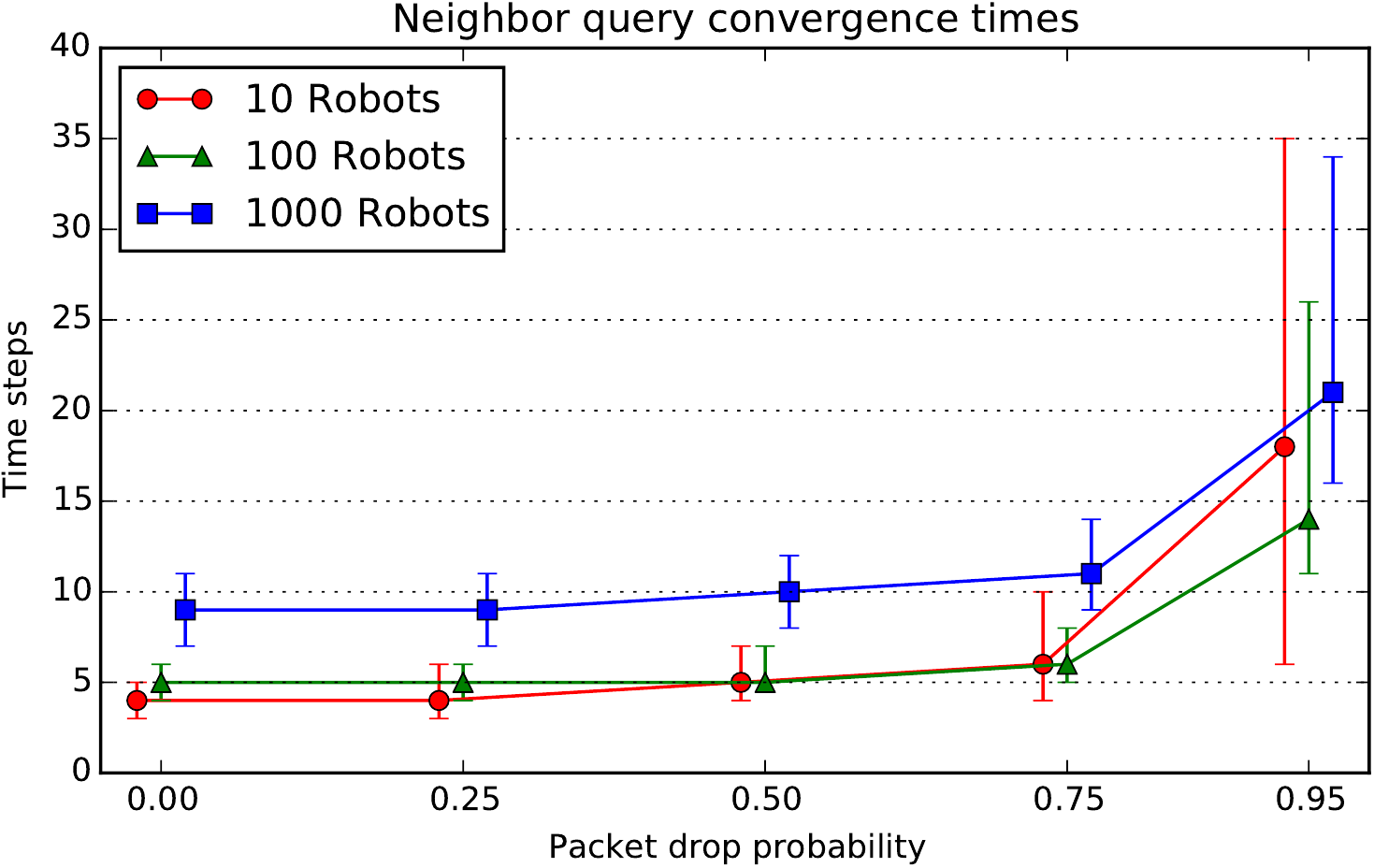}
  \begin{lstlisting}
# Constant for infinite distance (500 m)
DISTANCE_INF = 50000.

# Executed at init time
function init() {
  if(id == 0)
    # Source robot
    mydist = 0.
  else {
    # Other robots
    mydist = DISTANCE_INF
    # Listen to other robots' distances
    neighbors.listen("dist_to_source",
      function(vid, value, rid) {
        mydist = math.min(
          mydist,
          neighbors.get(rid).distance + value)
      })
  }
}

# Executed at each time step
function step() {
  neighbors.broadcast("dist_to_source", mydist)
}
  \end{lstlisting}
  \caption{Neighbor query performance assessment. The plot reports the
    median, max, and min values of the distributions obtained for each
    experimental configuration $\langle N, P\rangle$. The markers are
    slightly offset to make them visible.}
  \label{fig:nq_exp}
\end{figure}

\paragraph{Neighbor queries.}
To analyze the performance of neighbor queries, we devised an
experiment in which the robots must construct a distance gradient from
a robot acting as source. This experiment is representative because
the formation of gradients is a fundamental coordination mechanism in
swarm behaviors. As performance measure, we employ the time necessary
for every robot to estimate its distance to the source. The script and
the data plot are reported in \fig{fig:nq_exp}. The dynamics are
analogous to virtual stigmergy: convergence time depends weakly on $N$
and is unaffected by $P$ up to $P=0.75$; for
$\langle N=1000,P=0.75 \rangle$ convergence is reached in a maximum of
13 time steps; for $P=0.95$, convergence times increase sensibly.
Again, this is arguably due to the dense distribution of robots, which
facilitates the circulation of messages across the swarm, thus
mitigating the effects of message dropping.

\section{Related Work}
\label{sec:related}

Most of the literature in programming languages designed for robotics
focus on individual robots. Prominent examples of this body of work
are Willow Garage's Robot Operating System (ROS)~\cite{Quigley2009},
and the event-based language URBI~\cite{Baillie2005}.  In swarm
robotics, SWARMORPH-script~\cite{OGrady2012} is a bottom-up
behavior-based scripting language designed to achieve morphogenesis
with mobile robots.

The last decade saw the introduction of the first top-down approaches
to the development of distributed computing systems. Various
abstractions and programming languages have been proposed in the
sensor network community~\cite{Mottola2011}. A programming methodology
inspired by embryogenesis and designed for self-assembly applications
was proposed in~\cite{Nagpal2002}. Dantu \emph{et al.} proposed
Karma~\cite{Dantu2011}, a framework that combines centralized and
distributed elements to perform task allocation in a swarm of aerial
robots unable to communicate directly.

Proto~\cite{Bachrach2010} is a language designed for ``spatial
computers,'' that is, a collection of connected computing devices
scattered in a physical space. The spatial computer is modeled as a
continuous medium in which each point is assigned a tuple of
values. The primitive operations of Proto act on this medium. The
LISP-like syntax of Proto is modular by design and produces
predictable programs. Proto shines in scenarios in which homogeneous
devices perform distributed spatial computation---the inspiration for
Buzz' \texttt{neighbors} construct was taken from Proto. However, as a
language for robotics, Proto presents a number of limitations:
\begin{inparaenum}[\it (i)]
\item As a functional language, maintaining state over time is
  cumbersome;
\item Every node handles only single tuples;
\item Support for robot motion is limited;
\item No explicit support for heterogeneous devices is present.
\end{inparaenum}
Part of these issues have been addressed in~\cite{Viroli2012}.

Meld~\cite{Ashley-Rollman2009} is a declarative language that realizes
the top-down approach by allowing the developer to specify a
high-level, logic description of what the swarm as a whole should
achieve. The low-level (communication/coordination) mechanisms that
reify the high-level goals, i.e., the how, are left to the language
implementation and are transparent to the developer. The main concepts
of the language are facts and rules. A fact encodes a piece of
information that the system considers true at a given time. A
computation in Meld consists of applying the specified rules
progressively to produce all the true facts, until no further
production is possible. Meld supports heterogeneous robot swarms by
endowing each robot with facts that map to specific capabilities. A
similar concept exists in Buzz, with robot-specific symbols (see
\sect{sec:implcompilation}). The main limitation of Meld for swarm
robotics is the fact that its rule-based mechanics produce programs
whose execution is difficult to predict and debug, and it is thus
impossible to decompose complex programs into well-defined modules.

Voltron~\cite{Mottola2014} is a language designed for distributed
mobile sensing. Voltron allows the developer to specify the logic to
be executed at several locations, without having to dictate how the
robots must coordinate to achieve the objectives. Coordination is
achieved automatically through the use of a shared tuple space, for
which two implementations were tested---a centralized one, and a
decentralized one based on the concept of virtual
synchrony~\cite{Birman1987}. In Buzz, virtual stigmergy was loosely
inspired by the capabilities of virtual synchrony, although the
internals of the two systems differ substantially. Voltron excels in
single-robot, single-task scenarios in which pure sensing is involved;
however, fine-grained coordination of heterogeneous swarms is not
possible, because Voltron's abstraction hides the low-level details of
the robots.

\section{Conclusions and Future Work}
\label{sec:conclusions}
We presented Buzz, a novel programming language designed for
large-scale, heterogeneous robot swarms. The contributions of our work
include:
\begin{inparaenum}[\it (i)]
\item a mixed paradigm for the implementation of robot swarms, which
  allows the developer to specify fine-grained, bottom-up logic as
  well as reason in a top-down, swarm-oriented fashion;
\item the definition of a compositional and predictable approach to
  swarm behavior development;
\item the implementation of a general language capable of expressing
  the most common swarm behaviors.
\end{inparaenum}

Besides these contributions, we believe that one of the most important
aspects of Buzz is its potential to become an enabler for future
research on real-world, complex swarm robotics systems. Currently, no
standardized platform exists that allows researchers to compare,
share, and reuse swarm behaviors. Inescapably, development involves a
certain amount of re-coding of recurring swarm behaviors, such as
flocking, barriers, and creation of gradients. The design of Buzz is
motivated and nurtured by the necessity to overcome this state of
affairs. We hope that Buzz will have a lasting impact on the growth of
the swarm robotics field.

Future work on Buzz will involve several activities. Firstly, we will
integrate the run-time into multiple robotics platforms of different
kinds, such as ground-based and aerial robots. Secondly, we will
create a library of well-known swarm behaviors, which will be offered
open-source to practitioners as part of the Buzz distribution. This
will be the first true collection of `swarm patterns', in the
classical sense the word `pattern' assumes in software
engineering~\cite{Gamma1995}. Finally, we will tackle the design of
general approaches to swarm behavior debugging and fault
detection. These topics have received little attention in the
literature, and Buzz constitutes an ideal platform to study them. In
particular, we will study more in depth the impact of the network
topology on the efficiency of message passing, and we will investigate
adaptive methods to detect and mitigate the issues of unoptimal
topologies.

\section*{Acknowledgments}
\begin{footnotesize}
  \noindent This work was supported by an NSERC Engage
  Grant. Computations were made on the supercomputer Mammouth-Ms from
  Universit\'{e} de Sherbrooke, managed by Calcul Qu\'{e}bec and
  Compute Canada. The operation of this supercomputer is funded by the
  Canada Foundation for Innovation (CFI), NanoQu\'{e}bec, RMGA and the
  Fonds de recherche du Qu\'{e}bec --- Nature et technologies
  (FRQ-NT).
\end{footnotesize}

\bibliographystyle{IEEEtran}
\bibliography{tro2015}

\begin{thebibliography}{10}
\providecommand{\url}[1]{#1}
\csname url@samestyle\endcsname
\providecommand{\newblock}{\relax}
\providecommand{\bibinfo}[2]{#2}
\providecommand{\BIBentrySTDinterwordspacing}{\spaceskip=0pt\relax}
\providecommand{\BIBentryALTinterwordstretchfactor}{4}
\providecommand{\BIBentryALTinterwordspacing}{\spaceskip=\fontdimen2\font plus
\BIBentryALTinterwordstretchfactor\fontdimen3\font minus
  \fontdimen4\font\relax}
\providecommand{\BIBforeignlanguage}[2]{{%
\expandafter\ifx\csname l@#1\endcsname\relax
\typeout{** WARNING: IEEEtran.bst: No hyphenation pattern has been}%
\typeout{** loaded for the language `#1'. Using the pattern for}%
\typeout{** the default language instead.}%
\else
\language=\csname l@#1\endcsname
\fi
#2}}
\providecommand{\BIBdecl}{\relax}
\BIBdecl

\bibitem{Beni2005}
G.~Beni, ``{From Swarm Intelligence to Swarm Robotics},'' \emph{Swarm
  Robotics}, vol. 3342, pp. 1--9, 2005.

\bibitem{McLurkin2006}
J.~McLurkin, J.~Smith, J.~Frankel, D.~Sotkowitz, D.~Blau, and B.~Schmidt,
  ``{Speaking Swarmish : Human-Robot Interface Design for Large Swarms of
  Autonomous Mobile Robots},'' in \emph{AAAI Spring Symposium: To Boldly Go
  Where No Human-Robot Team Has Gone Before}.\hskip 1em plus 0.5em minus
  0.4em\relax AAAI, 2006, pp. 3--6.

\bibitem{Brambilla2013}
M.~Brambilla, E.~Ferrante, M.~Birattari, and M.~Dorigo, ``{Swarm robotics: a
  review from the swarm engineering perspective},'' \emph{Swarm Intelligence},
  vol.~7, no.~1, pp. 1--41, Jan. 2013.

\bibitem{Scott2006}
M.~Scott, \emph{Programming Language Pragmatics}.\hskip 1em plus 0.5em minus
  0.4em\relax San Francisco, CA: Morgan Kaufmann Publishers, 2006.

\bibitem{Stoy2001}
K.~St{\o}y, ``{Using situated communication in distributed autonomous mobile
  robots},'' in \emph{Proceedings of the 7th Scandinavian Conference on
  Artificial Intelligence}.\hskip 1em plus 0.5em minus 0.4em\relax IOS Press,
  2001, pp. 44--52.

\bibitem{Spears2004}
W.~M. Spears, D.~F. Spears, J.~C. Hamann, and R.~Heil, ``{Distributed,
  Physics-Based Control of Swarms of Vehicles},'' \emph{Autonomous Robots},
  vol.~17, no. 2/3, pp. 137--162, Sep. 2004.

\bibitem{Ferrante2014}
E.~Ferrante, A.~E. Turgut, A.~Stranieri, C.~Pinciroli, M.~Birattari, and
  M.~Dorigo, ``A self-adaptive communication strategy for flocking in
  stationary and non-stationary environments,'' \emph{Natural Computing},
  vol.~13, no.~2, pp. 225--245, 2014.

\bibitem{Pinciroli2015}
C.~Pinciroli, M.~Bonani, F.~Mondada, and M.~Dorigo, ``Adaptation and awareness
  in robot ensembles: Scenarios and algorithms,'' in \emph{Software Engineering
  for Collective Autonomic Systems}, M.~Wirsing, M.~H\"olzl, N.~Koch, and
  P.~Mayer, Eds.\hskip 1em plus 0.5em minus 0.4em\relax Springer International
  Publishing, 2015, vol. LNCS 8998, ch. IV.2, pp. 471--494.

\bibitem{Brutschy2014}
A.~Brutschy, G.~Pini, C.~Pinciroli, M.~Birattari, and M.~Dorigo,
  ``Self-organized task allocation to sequentially interdependent tasks in
  swarm robotics,'' \emph{Autonomous Agents and Multi-Agent Systems}, vol.~28,
  no.~1, pp. 101--125, 2014.

\bibitem{Rubenstein2012}
M.~Rubenstein, C.~Ahler, and R.~Nagpal, ``{Kilobot: A low cost scalable robot
  system for collective behaviors},'' \emph{2012 IEEE International Conference
  on Robotics and Automation}, pp. 3293--3298, May 2012.

\bibitem{Mondada2006}
F.~Mondada, M.~Bonani, X.~Raemy, J.~Pugh, C.~Cianci, A.~Klaptocz, J.-C.
  Zufferey, D.~Floreano, and A.~Martinoli, ``{The e-puck , a Robot Designed for
  Education in Engineering},'' in \emph{Proceedings of Robotica 2009 -- 9th
  Conference on Autonomous Robot Systems and Competitions}, P.~J.~S.
  Gon\c{c}alves, P.~J.~D. Torres, and C.~M.~O. Alves, Eds., vol.~1.\hskip 1em
  plus 0.5em minus 0.4em\relax IPCB, Castelo Branco, Portugal, 2006, pp.
  59--65.

\bibitem{Bonani2010}
M.~Bonani, V.~Longchamp, S.~Magnenat, P.~R\'etornaz, D.~Burnier, G.~Roulet,
  F.~Vaussard, H.~Bleuler, and F.~Mondada, ``The {marXbot}, a miniature mobile
  robot opening new perspectives for the collective-robotic research,'' in
  \emph{Proceedings of the {IEEE/RSJ} International Conference on Intelligent
  Robots and Systems ({IROS})}.\hskip 1em plus 0.5em minus 0.4em\relax
  Piscataway, NJ: {IEEE} Press, 2010, pp. 4187--4193.

\bibitem{DeSilva2014}
O.~De~Silva, G.~Mann, and R.~Gosine, ``An ultrasonic and vision-based relative
  positioning sensor for multirobot localization,'' \emph{{IEEE} Sensors
  Journal}, vol.~15, no.~3, pp. 1716--1726, 2014.

\bibitem{Jelasity2005}
M.~Jelasity, A.~Montresor, and B.~O., ``Gossip-based aggregation in large
  dynamic networks,'' \emph{ACM Transactions on Computer Systems}, vol.~23,
  no.~3, pp. 219--252, 2005.

\bibitem{Ierusalimschy2005}
R.~Ierusalimschy, L.~H. De~Figueiredo, and W.~Celes~Filho, ``The implementation
  of lua 5.0,'' \emph{Journal of Universal Computer Science}, vol.~11, no.~7,
  pp. 1159--1176, 2005.

\bibitem{Landin1964}
P.~J. Landin, ``The mechanical evaluation of expressions,'' \emph{The Computer
  Journal}, vol.~6, no.~4, pp. 308--320, 1964.

\bibitem{Gelernter1985}
D.~Gelernter, ``Generative communication in {Linda},'' \emph{{ACM Transactions
  on Programming Languages and Systems (TOPLAS)}}, vol.~7, no.~1, pp. 80--112,
  1985.

\bibitem{Grasse1959}
P.~Grass\'e, ``La reconstruction du nid et les coordinations
  inter-individuelles chez bellicositermes natalensis et cubitermes sp. la
  th\'eorie de la stigmergie: Essai d'interpr\'etation des termites
  constructeurs.'' \emph{Insects Sociaux}, vol.~6, pp. 41--83, 1959.

\bibitem{Voulgaris2005}
S.~Voulgaris, D.~Gavidia, and M.~{Van Steen}, ``{CYCLON: Inexpensive membership
  management for unstructured P2P overlays},'' \emph{Journal of Network and
  Systems Management}, vol.~13, no.~2, pp. 197--216, 2005.

\bibitem{Lamport1978}
L.~Lamport, ``Time, clocks, and the ordering of events in a distributed
  system,'' \emph{Communications of the ACM}, vol.~21, no.~7, pp. 558--565,
  Jul. 1978.

\bibitem{Quigley2009}
M.~Quigley, K.~Conley, B.~P. Gerkey, J.~Faust, T.~Foote, J.~Leibs, R.~Wheeler,
  and A.~Y. Ng, ``{ROS: an open-source Robot Operating System},'' in \emph{ICRA
  workshop on open source software}, 2009, p.~5.

\bibitem{Pinciroli2012}
C.~Pinciroli, V.~Trianni, R.~O'Grady, G.~Pini, A.~Brutschy, M.~Brambilla,
  N.~Mathews, E.~Ferrante, G.~{Di Caro}, F.~Ducatelle, M.~Birattari, L.~M.
  Gambardella, and M.~Dorigo, ``{ARGoS}: a modular, parallel, multi-engine
  simulator for multi-robot systems,'' \emph{Swarm Intelligence}, vol.~6,
  no.~4, pp. 271--295, 2012.

\bibitem{Dorigo2013}
M.~Dorigo, D.~Floreano, L.~Gambardella, F.~Mondada, S.~Nolfi, T.~Baaboura,
  M.~Birattari, M.~Bonani, M.~Brambilla, A.~Brutschy, D.~Burnier, A.~Campo,
  A.~Christensen, A.~Decugni\`{e}re, G.~{Di Caro}, F.~Ducatelle, E.~Ferrante,
  A.~F\"{o}rster, J.~Guzzi, V.~Longchamp, S.~Magnenat, J.~{Martinez Gonzales},
  N.~Mathews, M.~{Montes de Oca}, R.~O'Grady, C.~Pinciroli, G.~Pini,
  P.~R\'{e}tornaz, J.~Roberts, V.~Sperati, T.~Stirling, A.~Stranieri,
  T.~St\"{u}tzle, V.~Trianni, E.~Tuci, A.~Turgut, and F.~Vaussard,
  ``Swarmanoid: a novel concept for the study of heterogeneous robotic
  swarms,'' \emph{IEEE Robotics \& Automation Magazine}, vol.~20, no.~4, pp.
  60--71, 2013.

\bibitem{Nagpal2004}
R.~Nagpal, ``{A Catalog of Biologically-Inspired Primitives for Engineering
  Self-Organization},'' in \emph{Engineering Self-Organising Systems}, G.~{Di
  Marzo Serugendo}, A.~Karageorgos, O.~F. Rana, and F.~Zambonelli, Eds.\hskip
  1em plus 0.5em minus 0.4em\relax Springer Berlin Heidelberg, 2004, pp.
  53--62.

\bibitem{Baillie2005}
J.-C. Baillie, ``{URBI: towards a universal robotic low-level programming
  language},'' in \emph{2005 IEEE/RSJ International Conference on Intelligent
  Robots and Systems}.\hskip 1em plus 0.5em minus 0.4em\relax IEEE Press,
  Piscataway, NJ, 2005, pp. 820--825.

\bibitem{OGrady2012}
R.~O'Grady, A.~L. Christensen, and M.~Dorigo, ``{SWARMORPH: Morphogenesis with
  Self-Assembling Robots},'' in \emph{Morphogenetic Engineering: Toward
  Programmable Complex Systems}, R.~Doursat, H.~Sayama, and O.~Michel,
  Eds.\hskip 1em plus 0.5em minus 0.4em\relax Springer Berlin Heidelberg, 2012,
  ch.~2, pp. 27--60.

\bibitem{Mottola2011}
L.~Mottola and G.~P. Picco, ``{Programming wireless sensor networks:
  Fundamental concepts and state of the art},'' \emph{ACM Computing Surveys
  (CSUR)}, vol.~43, no.~3, p. Paper ID 19, 2011.

\bibitem{Nagpal2002}
R.~Nagpal, ``{Programmable self-assembly using biologically-inspired multiagent
  control},'' in \emph{Proceedings of the first international joint conference
  on Autonomous agents and multiagent systems part 1 - AAMAS '02}.\hskip 1em
  plus 0.5em minus 0.4em\relax New York, New York, USA: ACM Press, 2002, pp.
  418--425.

\bibitem{Dantu2011}
K.~Dantu, B.~Kate, J.~Waterman, P.~Bailis, and M.~Welsh, ``{Programming
  micro-aerial vehicle swarms with Karma},'' in \emph{Proceedings of the 9th
  ACM Conference on Embedded Networked Sensor Systems - SenSys '11}.\hskip 1em
  plus 0.5em minus 0.4em\relax New York, New York, USA: ACM Press, 2011, pp.
  121--134.

\bibitem{Bachrach2010}
J.~Bachrach, J.~Beal, and J.~McLurkin, ``{Composable continuous-space programs
  for robotic swarms},'' \emph{Neural Computing and Applications}, vol.~19,
  no.~6, pp. 825--847, May 2010.

\bibitem{Viroli2012}
M.~Viroli, D.~Pianini, and J.~Beal, ``{Linda in Space-Time: An Adaptive
  Coordination Model for Mobile Ad-Hoc Environments},'' \emph{Coordination
  2012, {LNCS} 7274}, pp. 212--229, 2012.

\bibitem{Ashley-Rollman2009}
M.~P. Ashley-Rollman, P.~Lee, S.~C. Goldstein, P.~Pillai, and J.~D. Campbell,
  ``{A language for large ensembles of independently executing nodes},'' in
  \emph{Logic Programming}, ser. {LNCS} 5649, P.~M. Hill and D.~S. Warren,
  Eds., vol. 5649 LNCS.\hskip 1em plus 0.5em minus 0.4em\relax Springer Berlin
  Heidelberg, 2009, pp. 265--280.

\bibitem{Mottola2014}
L.~Mottola, M.~Moretta, K.~Whitehouse, and C.~Ghezzi, ``{Team-level Programming
  of Drone Sensor Networks},'' in \emph{SenSys '14 Proceedings of the 12th ACM
  Conference on Embedded Network Sensor SystemsSystems}.\hskip 1em plus 0.5em
  minus 0.4em\relax ACM New York, NY, 2014, pp. 177--190.

\bibitem{Birman1987}
K.~Birman and T.~Joseph, ``Exploiting virtual synchrony in distributed
  systems,'' in \emph{{Proceedings of the eleventh ACM Symposium on Operating
  systems principles (SOSP '87)}}, vol.~21, no.~5.\hskip 1em plus 0.5em minus
  0.4em\relax ACM New York, 1987, pp. 123--138.

\bibitem{Gamma1995}
E.~Gamma, R.~Helm, R.~Johnson, and J.~Vlissides, \emph{Design Patterns:
  Elements of Reusable Object-Oriented Software}.\hskip 1em plus 0.5em minus
  0.4em\relax Addison-Wesley, 1995.

\end{thebibliography}

\end{document}